\documentclass[letterpaper]{article} 
\usepackage{aaai25}  
\usepackage{times}  
\usepackage{helvet}  
\usepackage{courier}  
\usepackage[hyphens]{url}  
\usepackage{graphicx} 
\usepackage{natbib}  
\usepackage{caption} 
\usepackage{algorithm}
\usepackage{algorithmic}
\usepackage[algo2e,ruled,boxed]{algorithm2e}
\usepackage{amsmath}
\usepackage{adjustbox}
\usepackage{tabularray}
\usepackage{multirow}
\usepackage{amsfonts,amssymb}
\usepackage{booktabs,makecell}
\usepackage{amsmath}
\usepackage{newfloat}
\usepackage{listings}
\DeclareCaptionStyle{ruled}{labelfont=normalfont,labelsep=colon,strut=off} 
\floatstyle{ruled}
\newfloat{listing}{tb}{lst}{}
\floatname{listing}{Listing}
\title{EGSRAL:An \underline{E}nhanced 3D \underline{G}aussian \underline{S}platting based \underline{R}enderer with \underline{A}utomated \underline{L}abeling for Large-Scale Driving Scene}
\author{
    Yixiong Huo\thanks{Equal contribution.},
    Guangfeng Jiang\footnotemark[1],
    Hongyang Wei,
    Ji Liu\thanks{Corresponding author.},
    Song Zhang,
    Han Liu,
    Xingliang Huang,
    Mingjie Lu,
    Jinzhang Peng,
    Dong Li,
    Lu Tian,
    Emad Barsoum
}
\affiliations{
    \textsuperscript{\rm 1}Advanced Micro Devices, Inc., Beijing, China \\

        \{yixiong.huo, guangfeng.jiang, ji.liu, mingjiel, jinz.peng, d.li,  lu.tian, ashish.sirasao\}@amd.com
}

\usepackage{bibentry}

\begin{document}

\maketitle

\begin{abstract}
    3D Gaussian Splatting (3D GS) has gained popularity due to its faster rendering speed and high-quality novel view synthesis. Some researchers have explored using 3D GS for reconstructing driving scenes. However, these methods often rely on various data types, such as depth maps, 3D boxes, and trajectories of moving objects. Additionally, the lack of annotations for synthesized images limits their direct application in downstream tasks.
    To address these issues, we propose EGSRAL, a 3D GS-based method that relies solely on training images without extra annotations. EGSRAL enhances 3D GS's capability to model both dynamic objects and static backgrounds and introduces a novel adaptor for auto labeling, generating corresponding annotations based on existing annotations. We also propose a grouping strategy for vanilla 3D GS to address perspective issues in rendering large-scale, complex scenes. Our method achieves state-of-the-art performance on multiple datasets without any extra annotation. For example, the PSNR metric reaches 29.04 on the nuScenes dataset. Moreover, our automated labeling can significantly improve the performance of 2D/3D detection tasks. Code is available at https://github.com/jiangxb98/EGSRAL.

\end{abstract}

%

\section{Introduction}
    Synthesizing novel photorealistic views represents a complex and crucial challenge in the fields of computer vision and graphics. With the rapid development of neural radiance fields~(NeRFs) \cite{nerf}, free-view synthesis has gradually transferred to the domain of large-scale view synthesis, particularly in synthesizing streetscapes critical for autonomous driving \cite{multimedia}. However, simulating outdoor environments is challenging due to the complexity of geographical locations, diverse surroundings, and varying road conditions. Image-to-image translation methods~\cite{wang2018high,isola2017image} have been proposed to synthesize semantically labeled streetscapes by learning the mapping between source and target images. While these methods generate visually impressive street views, they often exhibit noticeable artifacts and inconsistent textures in local details. Additionally, the relatively uniform viewpoints of the synthesized images present challenges for their application in complex autonomous driving scenes.

    To address these challenges, Drive-3DAu~\cite{3DAU} introduces a 3D data augmentation approach using NeRF, designed to augment driving scenes in the 3D space. DGNR~\cite{DGNR} presents a novel framework that learns a density space from scenes to guide the construction of a point-based renderer. Meanwhile, READ~\cite{read} offers a large-scale driving simulation environment for generating realistic data for advanced driver assistance systems. 3D GS-based methods~\cite{zhou2024drivinggaussian,yan2024street} have been employed to synthesize driving scenes due to their superior generation capabilities.
    While these methods generate realistic images suitable for autonomous driving, they cannot simultaneously synthesize novel views and provide corresponding 2D/3D annotation boxes, which are crucial for supervised model training. \textbf{Consequently, enhancing novel view synthesis for large-scale scenes and achieving automatic annotations for new views remain key challenges in autonomous driving.}

    To overcome these challenges, we introduce a novel framework called EGSRAL based on an enhanced 3D Gaussian Splatting (3D GS) technique. This framework improves the quality of novel view synthesis while simultaneously generating corresponding annotations. Specifically, we propose a deformation enhancement module to refine the Gaussian deformation field enhancing the modeling of both dynamic objects and static backgrounds. Additionally, we introduce an opacity enhancement module that leverages a neural network, replacing the original learnable parameter, to significantly boost the modeling capacity of complex driving scenes. Furthermore, to address the issue of unrealistic viewpoints in rendering large-scale complex scenes, where occluded distant Gaussians should not be included, we also propose a grouping strategy for vanilla 3D GS.

    In short, our contributions are as follows: 
    (1)~We propose an enhanced 3D GS-based renderer called EGSRAL, which can synthesize novel view images with corresponding annotations based on existing dataset annotations. EGSRAL introduces a deformation enhancement module and an opacity enhancement module, both of which enhance the modeling capability of 3D GS for complex scenes.
    (2)~Additionally, to address the issue of unreasonable perspectives in rendering large-scale, complex scenes, we introduce a grouping strategy for vanilla 3D GS.
    (3)~Unlike previous methods that focus solely on novel view synthesis, we propose an adaptor with three constraints to transform neighboring annotation boxes into novel annotation boxes in the domain of autonomous driving.
    (4)~ Experimental results demonstrate that our method outperforms existing rendering methods for large-scale scenes. Additionally, the novel view images with corresponding annotations effectively improve the performance of 2D/3D detection models.

\section{Related Work}
\subsection{3D Gaussian Splatting}
    Recently 3D GS has achieved exciting results in novel view synthesis and real-time rendering. Unlike NeRF, which uses implicit functions and volumetric rendering, 3D GS leverages a set of 3D Gaussians to model the scene, allowing for fast rendering through a tile-based rasterizer.
    
    The original 3D GS \cite{3dgs} is designed to model static scenes, and subsequent work has tried to extend it to dynamic scenes. \cite{d3dgs} introduce a deformation field to model monocular dynamic scenes, which use encoded Gaussian position and timestep to predict the position, rotation, and scaling offsets of 3D Gaussian. 4D GS \cite{wu20244d} uses HexPlane \cite{cao2023hexplane} to encode temporal and spatial information for 3D Gaussian, building Gaussian features from 4D neural voxels, and then predicting the deformation of 3D Gaussians. 
    However, compared to indoor scenes, outdoor scenes present greater challenges due to their larger spatial extent and the wider range of motion of dynamic objects. The above 3D GS methods often do not scale well to large-scale outdoor autonomous driving scenes. In contrast, our approach is more effective at reconstructing autonomous driving scenes.
    
\subsection{Novel-View Synthesis for Driving Scene}
    Autonomous driving scene reconstruction methods based on NeRF can be roughly divided into two categories: perception-based \cite{panoptic-nerf, nerf-det, panoptic-nerf-field, in-place} and simulation-based \cite{mars, nerf-lidar, read}. Perception-based methods leverage capabilities of NeRF to capture the semantic information and geometrical representations. For simulation, MARS \cite{mars} models the foreground objects and background environments separately based on NeRF, making it flexible for scene controlling in autonomous driving simulation. Furthermore, READ \cite{read} explores diverse sampling strategies aimed at enabling the synthesis of expansive driving scenarios.
    
    Some recent works have begun to bring 3D GS into the reconstruction of driving scenes. DrivingGaussian \cite{zhou2024drivinggaussian} uses incremental static Gaussians and composite dynamic Gaussians to reconstruct static scenes and dynamic objects, respectively. Street Gaussians \cite{yan2024street} represents the scene as a set of point-based background and foreground objects where an optimization strategy is introduced to deal with dynamic foreground vehicles. 
    Although the above methods introduce a novel view synthesis paradigm into the autonomous driving domain, mainstream methods still heavily rely on annotated data. Simply synthesizing novel views without annotations can not be directly applied to the downstream perception tasks. Therefore, to address this challenge, we first propose a new framework for synthesizing new views based on the 3D GS while enabling automatic annotation of downstream perception tasks.

\begin{figure*}[!t]
    \centering 
    \includegraphics[width=1.0\linewidth]{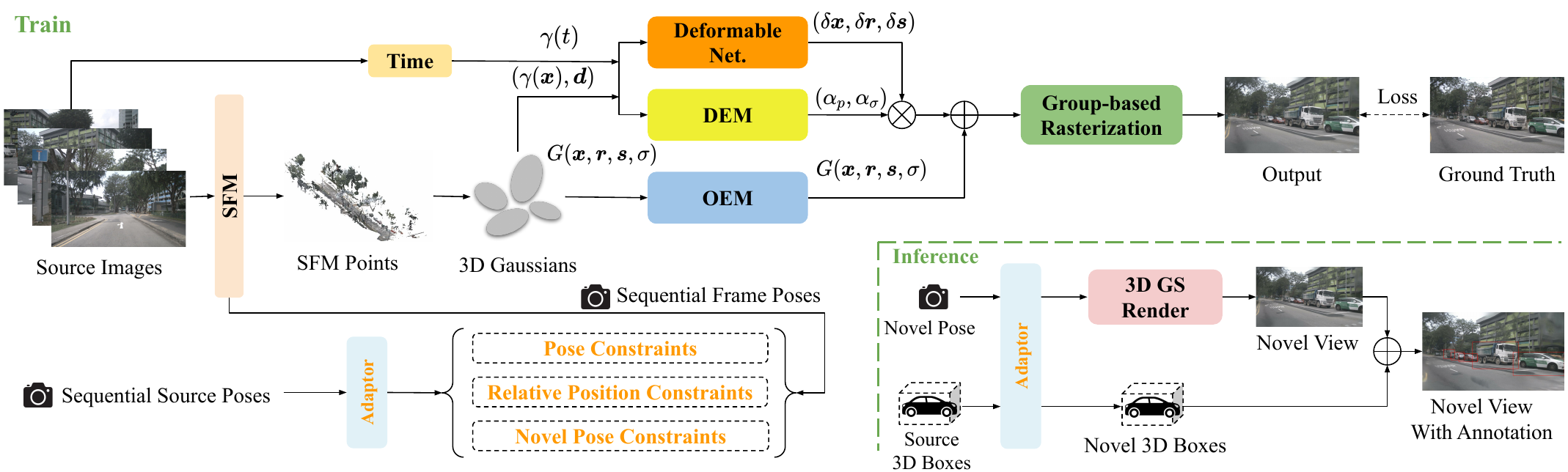}
    \caption{Illustration of overall EGSRAL. The EGSRAL framework begins by aligning the input image, followed by initializing the 3D Gaussian using the point cloud generated by the SfM. A deformable network (orange block) constructs the 3D Gaussian deformation field while the deformation enhancement module (DEM) (yellow blocks) refines this field. The opacity enhancement module (OEM) (blue blocks) optimizes opacity. To address perspective issues in large-scale, complex scenes, a group-based training and rendering strategy (green block) is employed (Section 3.2). Additionally, the adaptor is trained using three constraints (orange) to enhance its coordinate relationship modeling. During inference, these modules render and synthesize novel view images with corresponding annotations (Section 3.3).}
    \label{fig:figure_1} 
\end{figure*}

\subsection{Auto Labeling With Scene Reconstruction}
    Currently, some works partially explored the potential of NeRFs to serve as data factories at a level to achieve auto labeling in other task domains ~(e.g., depth estimation \cite{rc-mvsnet}, object detection \cite{nerual-sim}, semantic segmentation \cite{in-place}) to further improve domain performance. For instance, NS \cite{ns} introduces a NeRF-supervised training procedure which exploits rendered stereo triplets to compensate for occlusions and depth maps as proxy labels to improve the depth stereo performance. Semantic-NeRF \cite{in-place} extends NeRF to jointly encode semantics with appearance and geometry to complete and accurate 2D semantic labels which can be achieved using a small amount of in-place annotations specific to the scene. Neural-Sim \cite{nerual-sim} proposes a novel bilevel optimization algorithm to automatically optimize rendering parameters (pose, zoom, illumination) to generate optimal data for downstream tasks using NeRF, which substitutes the traditional graphics pipeline and synthesizes useful images. However, there is currently no work to introduce auto labeling into 3D GS scene reconstruction. Inspired by the above methods, we consider incorporating auto labeling in the autonomous driving domain to further advance downstream perception tasks.

\section{Methodology}

\subsection{Overview}
   Given input image sequences of a driving scene and point clouds estimated through structure-from-motion (SfM) methods~\cite{metashpe, schonberger2016structure}, our EGSRAL framework can synthesize realistic driving scenes from multiple perspectives while automatically labeling the corresponding novel synthetic views. We also propose a grouping strategy to address perspective issues in large-scale driving scenes. Our framework is divided into two parts: an Enhanced 3D GS Rendering and a Novel View Auto Labeling, as illustrated in Figure~\ref{fig:figure_1}. The 3D GS rendering is based on the Deformable 3D GS~\cite{d3dgs}, which we have extended with innovative modules~(Sec. \ref{sec:3.2}) to improve novel view synthesis. For auto labeling, we introduce an adaptor to transform camera poses and bounding boxes, generating corresponding annotations for the novel views~(Sec. \ref{sec:3.3}).

\subsection{Enhanced 3D GS Rendering}
    \label{sec:3.2}
    3D GS is typically used for modeling static scenes and its ability to handle dynamic scenes is limited. Recent research~\cite{d3dgs, sun20243dgstream, wu20244d, cao2023hexplane} has focused on improving dynamic scene modeling. We use Deformable 3D GS~\cite{d3dgs} as our baseline and apply it to automatic driving scene reconstruction. This approach employs a unified deformation field for both static backgrounds and dynamic objects. To better represent each Gaussian primitive's state, we introduce a state attribute $\boldsymbol{d}\in R^{d \times 1}$ which implicitly indicates whether the primitive is static or dynamic.
    We further enhance the deformation field by incorporating a deformation enhancement module. This module decodes each Gaussian primitive's state attribute $\boldsymbol{d}$ and time encoding $\gamma(t)$ to determine the adjustment factor for the deformation field. Recognizing the importance of opacity in rendering, we also introduce an opacity enhancement module to improve the capacity.

\noindent \textbf{Deformable 3D GS Network.}
    To reduce data dependency, we use only image data for driving scene reconstruction. We begin by initializing a set of 3D Gaussians $G_0 \left(\boldsymbol{x}, \boldsymbol{r}, \boldsymbol{s}, \sigma \right)$ through SfM, where $\boldsymbol{x}$, $\boldsymbol{r}$, $\boldsymbol{s}$, and $\sigma$ represent the position of the Gaussian primitive, quaternion, scaling, and opacity, respectively. To better model dynamic 3D Gaussians, the Deformable 3D GS introduces a deformation field for the 3D Gaussians. The deformation field takes the position $\boldsymbol{x}$ and time $t$ as inputs to predict offsets $\delta\boldsymbol{x}$, $\delta\boldsymbol{r}$, and $\delta\boldsymbol{s}$ for $\boldsymbol{x}$, $\boldsymbol{r}$, and $\boldsymbol{s}$, respectively. Subsequently, these deformed 3D Gaussians $G_1 \left(\boldsymbol{x} + \delta\boldsymbol{x}, \boldsymbol{r} + \delta\boldsymbol{r}, \boldsymbol{s} + \delta\boldsymbol{s}, \sigma \right)$ are then passed through a differentiable tile rasterizer~\cite{3dgs} to render the novel image:
    \begin{equation}
    \label{eq: render equation}
    \begin{aligned}
        C_{\{u,v\}} &= \sum_{i \in N} T_i \alpha_i c_i
    \end{aligned}
    \end{equation}
    where $C_{\{u,v\}}$ denotes the rendered color of the $(u,v)$ pixel, $T_i$ is the transmittance defined by $\Pi_{j=1}^{i - 1}(1 - \alpha_{j}) $, $c_i$ is the color of each Gaussian primitive, and $\alpha_i$ is calculated by evaluating a 2D Gaussian with covariance matrix \cite{yifan2019differentiable} multiplied by a learned per-primitive opacity.
    
   However, Deformable 3D GS struggles to effectively model dynamic objects and static backgrounds. We believe this limitation stems from the inability of 3D Gaussians to adequately represent both dynamic and static elements. To address this, we introduce a state attribute $\boldsymbol{d}$ for each Gaussian primitive and incorporate it as an input to the deformable network. This enhancement improves the network’s ability to capture the state of the Gaussian primitives:
    \begin{equation}
    \label{eq: deform equation}
    \begin{aligned}
         \left( \delta\boldsymbol{x}, \delta\boldsymbol{r}, \delta\boldsymbol{s} \right) = \mathcal{F_\theta}(\gamma(\operatorname{sg}(\boldsymbol{x})), \gamma(t), \boldsymbol{d})
    \end{aligned}
    \end{equation}
    where $\mathcal{F_\theta}$ represents the deformable network's parameters, $\operatorname{sg}(\cdot)$  denotes a stop-gradient operation and $\gamma$ is the positional encoding.

\noindent \textbf{Deformation Enhancement Module (DEM).}
    To further enhance the modeling of dynamic objects, we propose a deformation enhancement module that fine-tunes the deformation field based on the time $t$ and the state attribute $\boldsymbol{d}$ of the Gaussian primitive. Specifically, we use the state attribute $\boldsymbol{d}$ and time encoding $\gamma(t)$  as inputs to output an adjustment factor $\alpha_{p}$ for each Gaussian primitive's deformation field:   
    \begin{equation}
    \label{eq: position factor}
    \begin{aligned}
        \alpha_{p} = sigmoid \left( MLP_1 \left(\mathcal{F_{\beta}}\left( \boldsymbol{d}, \gamma\left(t \right),\boldsymbol{x}-\boldsymbol{c}_{view} \right)\right)\right)
    \end{aligned}
    \end{equation}
    where $\mathcal{F_\beta}$ represents the dynamic encoding network's parameters, $\boldsymbol{c}_{view}$ is the current camera's coordinate position, and $sigmoid(\cdot)$ is the activation function.
    
    Additionally, in driving scenes, we dynamically adjust the opacity of dynamic objects using an opacity adjustment factor $\alpha_{\sigma}$ to more effectively model their appearance and disappearance:
    \begin{equation}
    \label{eq: opacity factor}
    \begin{aligned}
        \alpha_{\sigma} = tanh \left( MLP_2 \left(\mathcal{F_{\beta}}\left( \boldsymbol{d}, \gamma\left(t \right),\boldsymbol{x}-\boldsymbol{c}_{view} \right)\right)\right)
    \end{aligned}
    \end{equation}
    where $tanh(\cdot)$ is the activation function.
    
    Using the adjustment factors $\alpha_{p}$ and $\alpha_{\sigma}$, we modify the attributes of deformed 3D Gaussians $G_1$ to $G_2 \left(\boldsymbol{x} + \alpha_{p} \delta\boldsymbol{x}, \boldsymbol{r} + \delta\boldsymbol{r}, \boldsymbol{s} + \delta\boldsymbol{s}, \alpha_{\sigma}\sigma \right)$, enabling the deformation field to model dynamic objects and static backgrounds with greater detail.

\noindent \textbf{Opacity Enhancement Module (OEM).}
  As seen in the image rendering formula (Equation \ref{eq: render equation}), the rendering of the current pixel depends on the color and opacity of the Gaussian primitive. To enhance the capacity of opacity prediction, we initialize the Gaussian primitive by initializing opacity to a trainable parameter $\boldsymbol{\sigma}^{'} \in R^{16\times 1}$ and introduce a lightweight network to accelerate opacity optimization:
    \begin{equation}
    \label{eq: opacity up}
    \begin{aligned}
        \sigma = \left(\mathcal{F_{\eta}}\left(\boldsymbol{\sigma}^{'}\right)\right)
    \end{aligned}
    \end{equation}
    where $\mathcal{F_\eta}$ is the opacity enhancement module's parameters.
    \begin{figure}[t]
    \centering 
    \includegraphics[width=1.0\linewidth]{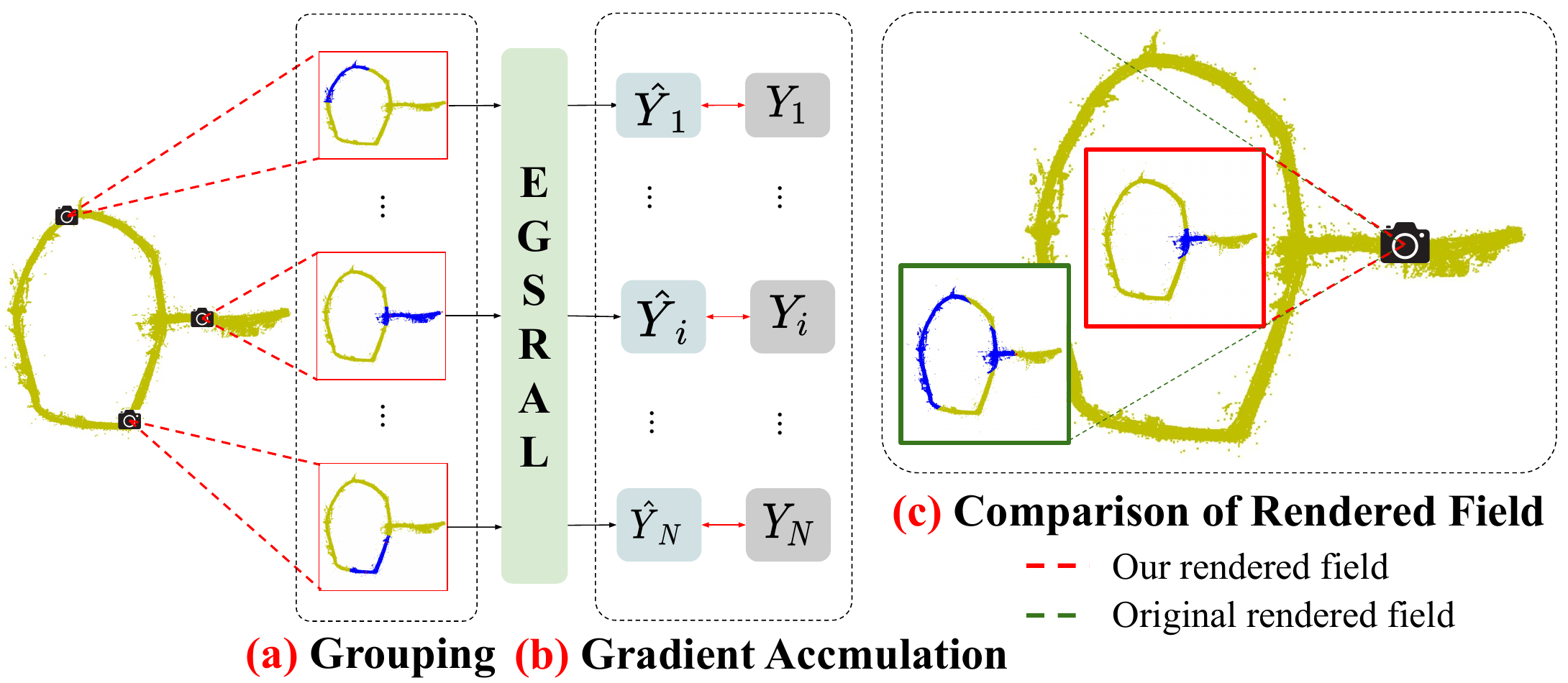}
    \caption{Illustration of the grouping strategy.}
    \label{fig: gps} 
    \end{figure}

\noindent \textbf{Grouping Strategy (GPS).}
    Previous methods \cite{zhou2024drivinggaussian, yan2024street} for reconstructing driving scenes using 3D GS have not accounted for large-scale scenes. To address this, we propose a grouping strategy. As illustrated in Figure \ref{fig: gps}(a), we divide the scene into $N$ groups using a fixed image interval and assign each Gaussian primitive a group identification $id$, which allows us to perform subsequent cloning, splitting, and rendering based on these group $id$s. By grouping strategy, we solve the problem of unreasonable fields of view. As shown in Figure \ref{fig: gps}(c), the original 3D GS rendering field of view includes distant Gaussian primitives (green rectangle), which is impractical. By grouping (red rectangle), we limit the field of view to a specific range, thus reducing the optimization burden by excluding Gaussian primitives outside this range.

    Unlike the static incremental training mode in DrivingGaussian \cite{zhou2024drivinggaussian}, our method avoids sequential group training and the need for separate deformable networks for each group, leading to fewer models and shorter training times. However, the varying positional distribution of Gaussian primitives can affect convergence. To address this, we introduce a multi-group joint optimization strategy. As shown in Figure~\ref{fig: gps}(b), each group performs forward propagation to accumulate gradients independently. After all groups have completed their forward passes, we perform gradient backpropagation to optimize the network parameters, stabilizing the training of the deformable network.

    Additionally, to address the poor reconstruction quality of 3D GS in the initial frames, we propose an overlap training strategy. Specifically, for each group, we use $N$ images from the previous group to train the current group, significantly improving reconstruction quality. The detailed algorithm is provided in Appendix Algorithm 1.
     
\subsection{Novel View Auto Labeling}
    \label{sec:3.3}
    \noindent \textbf{Adaptor Requirement}. Our method leverages image sequences (dataset) to construct scene point clouds and estimate camera poses and parameters using SfM methods. The generated point clouds and camera poses are defined in a new coordinate system generated by the SfM methods. However, the 3D annotations of the image sequences and corresponding camera poses are defined in the original world coordinate system, as in nuScenes~\cite{nuscenes}. Consequently, there exist two different coordinate systems: the original world coordinate system (OWCS) and the estimated world coordinate system (EWCS) from SfM methods.
    
    Our renderer is trained based on the EWCS, as it takes the estimated points and camera poses in the EWCS as input. The novel camera poses used for generating novel view images are also based on the EWCS, which complicates the utilization of the dataset's 3D annotations in the OWCS. A transformation adaptor is necessary to establish the relationship between the two coordinate systems, allowing us to utilize the 3D annotations effectively. This adaptor transforms coordinates from the OWCS to the EWCS. As a result, novel view camera poses from the OWCS can be converted into the EWCS and fed into the renderer to generate novel view images.

\begin{figure}[!t] 
        \centering 
        \includegraphics[width=0.85\linewidth]{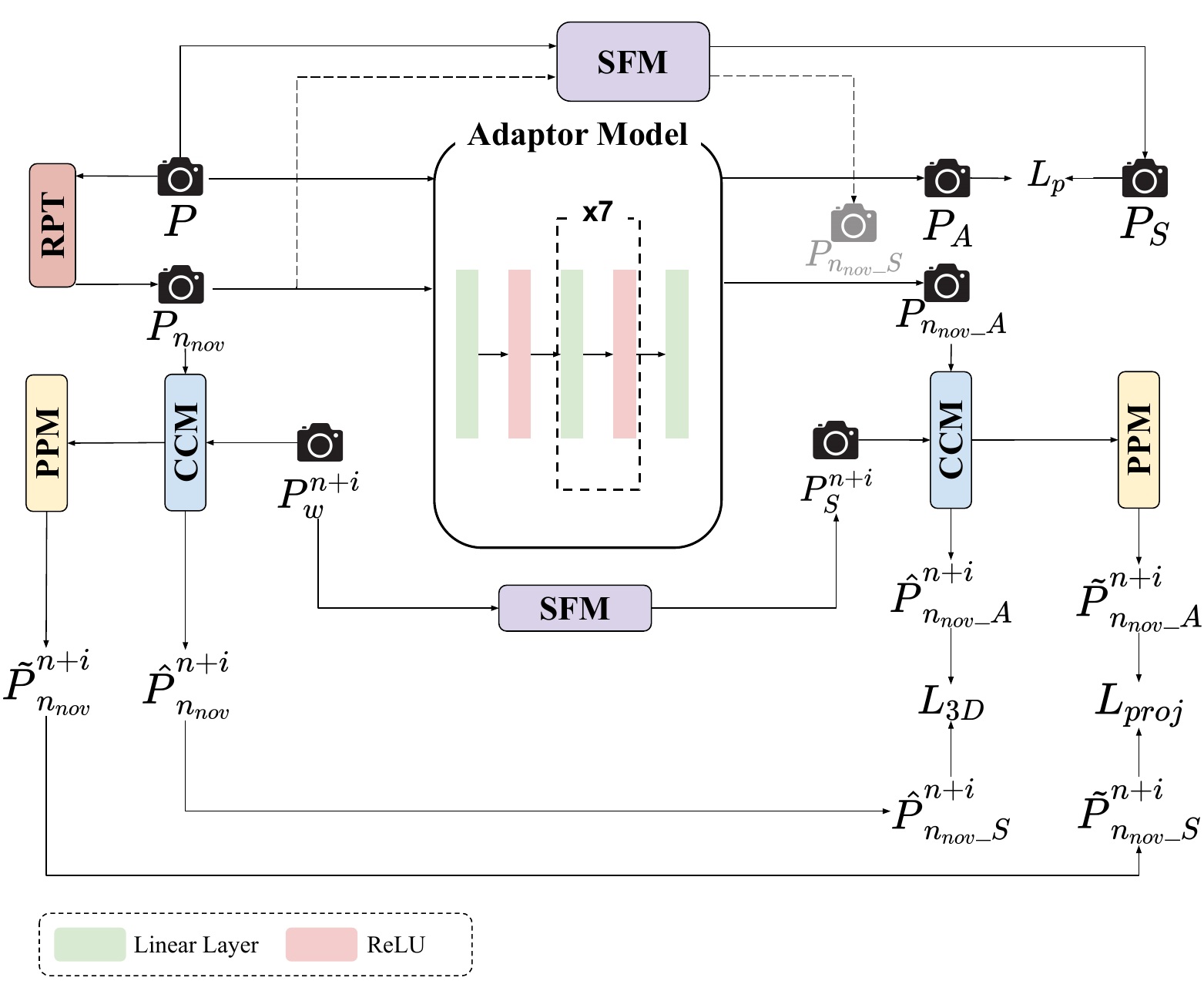}
        \caption{Illustration of the adaptor module including model and transformation modules. Camera pose $P_{n_{nov}\_S}$ in SfM coordinate system is the corresponding pose of novel camera pose $P_{n_{nov}}$.}
        \label{fig:Adaptor_mlp} 
    \end{figure}

    \noindent \textbf{Our Adaptor}.
    We take a neural network to model the transformation relationship of camera poses from the two coordinate systems, which takes camera poses in the OWCS as input and predicts the corresponding camera poses in the EWCS. We represent the camera pose as a 3$\times$4 matrix that includes both rotation and translation information. Therefore the shapes of input and output are both 3$\times$4. We take MLPs to build our adaptor, as shown in Figure~\ref{fig:Adaptor_mlp}, the backbone of our adaptor is composed of 8 layers of MLP, and the output head is a simple linear layer, which is used to predict camera poses in the EWCS. To optimize our adaptor, we introduce three constraints during the training process.


    \noindent \textbf{Constrains for Adaptor}.
   To effectively train the adaptor, we need camera poses from the dataset in both coordinate systems. It is straightforward to obtain the existing camera pose pairs from these systems, where each pair consists of camera poses of the same frame from both the OWCS and the EWCS. We adopt the  $smooth_{L_{1}}$ loss \cite{girshick2015fast} to ensure that the adaptor's predictions closely match the corresponding SfM camera poses in the EWCS, given the original camera poses in the OWCS.
    The loss function for camera pose constraints is:
     \begin{equation}
     \label{eq:equation_3}
     \begin{aligned}
         \ L_{p} = smooth_{L_{1}}(P_{A}, P_{S}) \\
     \end{aligned}
     \end{equation}
    \noindent where $L_{p}$ is the loss of the existing camera pose constraint, $P_{A}$ is the pose predicted by the adaptor given camera pose $P$ in the OWCS, and $P_{S}$ is the corresponding pose generated by SfM in the EWCS. After applying the initial constraint, the adaptor can transform existing camera poses between the two coordinate systems. However, due to the limited number of camera pose pairs, the adaptor's generalization ability is restricted, resulting in poor performance with novel camera poses not included in the existing pairs. To address this issue, we introduce new camera pose constraints.

    The novel camera pose constraints aim to enhance the adaptor's generalization ability by utilizing existing data to constrain new camera poses. Based on the projection consistency rule of the same objects in two camera coordinate systems, we can construct projection constraints for new camera poses. Specifically, for a camera pose $P$ in the dataset, we use a Random Position Transformation (RPT) module to sample a nearby novel pose $P_{nov}$. This pose has a corresponding camera pose $P_{nov\_S} $ in the EWCS, which is implicit and not directly available. Next, we take the following $N$ camera pose pairs after the current camera pose and project them onto the corresponding novel camera pose planes in both coordinate systems. The corresponding projected points should have the same pixel coordinates. As shown in  Figure~\ref{fig:Adaptor_mlp}, the coordinate conversion module (CCM) is used to convert points (the $N$ camera poses) in the OWCS to the camera coordinate system firstly by leveraging coordinate transformation formulas shown in Equation \ref{eq:CCM}. In the coordinate system of the novel camera pose, the relative pose of $ P_{n+i}$ is represented as $\hat P_{n_{nov}}^{n+i}$.
     \begin{equation}
     \label{eq:CCM}
    \begin{aligned}
        \ \hat P_{n_{nov}}^{n+i} = (P_{w}^{n_{nov}})^{-1} * P_{w}^{n+i}  \\
    \end{aligned}
     \end{equation}
    \noindent where $P_{w}^{n+i}$ is the camera pose of the $frame_{n+i}$ in the OWCS,  $P_{w}^{n_{nov}}$ is the novel camera pose, sampled nearby the current frame $frame_{n}$ in the OWCS. 
    Then, we project these points to the novel camera pose plane. As shown in  Figure~\ref{fig:Adaptor_mlp}, the point project module (PPM) is used to project points from the camera coordinate system to the pixel coordinate system. Next, we calculate the coordinate of $\hat P_{n_{nov}}^{n+i}$ in the pixel coordinate system of the novel camera pose plane by utilizing the camera intrinsic parameters, as illustrated in Equation \ref{eq:PPM_1} and \ref{eq:PPM_2} below:
    \begin{align}
    \label{eq:PPM_1}
    \acute{P}_{n_{nov}}^{n+i} &= K_{intr} *
    \begin{bmatrix} 
    \begin{aligned}
        T_{\hat P_{n_{nov}}^{n+i}}, \\
        \centering 1 
    \end{aligned}
    \end{bmatrix} \\
    \label{eq:PPM_2}
    \tilde{P}_{n_{nov}}^{n+i} &= 
    \begin{bmatrix} 
    \begin{aligned}
        x_{\acute{P}_{n_{nov}}^{n+i}} / z_{\acute{P}_{n_{nov}}^{n+i}}, \\
        y_{\acute{P}_{n_{nov}}^{n+i}} / z_{\acute{P}_{n_{nov}}^{n+i}},
    \end{aligned}
    \end{bmatrix}
    \end{align}

   \noindent where $K_{intr}$ is camera intrinsic parameter matrix, and $T_{\hat P_{n_{nov}}^{n+i}}$ is the translation of $\hat P_{n_{nov}}^{n+i}$. Then the projection constraint can be summarized as shown in Equation \ref{eq:loss_proj}.
    
    \begin{equation}
    \label{eq:loss_proj}
    \begin{aligned}
        \ L_{proj} = \sum_{i=1}^{N} MSE(\tilde{P}_{{n_{nov}}\_A}^{n+i}, \tilde{P}_{{n_{nov}}\_S}^{n+i})
    \end{aligned}
    \end{equation}
    
     \noindent where $\tilde{P}_{{n_{nov}}\_A}^{n+i}$ is the pixel coordinate of the relative camera pose $\hat P_{n_{nov}\_A}^{n+i}$ in the EWCS projected to the novel camera pose $P_{n_{nov}\_A}$ plane. $P_{n_{nov}\_A}$ is predicted by our adaptor fed with novel camera pose $P_{n_{nov}}$ as shown in Figure~\ref{fig:Adaptor_mlp}. Our adaptor aims to ensure that $P_{n_{nov}\_A}$ is equal to $P_{n_{nov}\_S}$ by making $\tilde{P}_{{n_{nov}}\_A}^{n+i}$ equal to $\tilde{P}_{{n_{nov}}\_S}^{n+i}$.
     $\tilde{P}_{{n_{nov}}\_S}^{n+i}$ is estimated by using the projected pixel coordinates from the corresponding camera poses in the OWCS as shown in Figure~\ref{fig:Adaptor_mlp} based on the projection consistency rule.
    
    
    Due to the limitations of pixel coordinate constraints (as in Equation \ref{eq:loss_proj}), which only restrict the relationship between $x$ and $z$ and between $y$ and $z$ without determining their specific values, the adaptor fails to converge to the desired outcome. Including the constraints on the position in the camera coordinate system (as in Equation \ref{eq:loss_3d}) into $L_{All}$ effectively resolves the aforementioned issue. Given the similarity between real camera intrinsic parameters and those estimated by SfM, it is possible to assume identical position information in the camera coordinate system during the initial stages of adaptor training. Subsequently, the weight on the camera coordinate system constraints can be appropriately reduced. The overall loss is expressed as in Equation \ref{eq:loss_all} below:
    \begin{align}
    \label{eq:loss_3d}
    L_{3D} &= \sum_{i=1}^{N} smooth_{L_{1}}(\hat{P}_{{n_{nov}}\_A}^{n+i}, \hat{P}_{{n_{nov}}\_S}^{n+i}) \\
    \label{eq:loss_all}
    L_{all} &= w_{1} * L_{p} + w_{2} * L_{3D} + w_{3} * L_{proj}
    \end{align}

    \noindent where $\hat{P}_{{n_{nov}}\_A}^{n+i}$ is the 3D coordinate of the relative camera pose of camera pose $ P_{S}^{n+i}$ in the EWCS relative to novel camera pose $P_{n_{nov}\_A}$, as shown in Figure~\ref{fig:Adaptor_mlp}. $\hat{P}_{{n_{nov}}\_S}^{n+i}$ is the estimated 3D coordinate of the relative camera pose of the camera pose $P_{S}^{n+i}$ in the EWCS, relative to novel camera pose $P_{n_{nov}\_S}$, obtained using the 3D coordinates from the corresponding camera poses in the OWCS. $ L_{3D}$ is the loss of novel camera pose constraints in 3D coordinates, $L_{proj}$ is the loss of novel camera pose constraints in project pixel coordinates, $w_{1}$ is the weight of $ L_{p}$, $w_{2}$ is the weight of $ L_{3D}$, $w_{3}$ is the weight of $ L_{proj}$.

    \noindent \textbf{Annotation Generation.}
    During the inference phase, as illustrated in Figure~\ref{fig:figure_1}, we start with the original camera pose $P_{ori}$ in the OWCS from the dataset. This pose undergoes an affine random transformation to obtain $P_{nov}$ by the RPT module shown in Figure~\ref{fig:Adaptor_mlp}. Then we feed $P_{nov}$ into the adaptor to produce the corresponding pose in the EWCS, denoted as $P_{nov\_s}$. Feeding $P_{nov\_s}$ into the renderer generates the corresponding novel view image, called $image_{nov}$. Similarly, applying the same affine transformation to the original dataset's annotations $anns_{ori}$ results in $anns_{nov}$, which corresponds to $image_{nov}$. Combining $image_{nov}$ with $anns_{nov}$ results in a novel view image with annotations, effectively accomplishing auto labeling.

\section{Experiment}
\subsection{Datasets}
    \noindent \textbf{KITTI Dataset.} KITTI dataset \cite{kitti} contains a wide range of driving scenes. Our experiments focus on a single distinct scene: City, which consists of 1424 train images and 159 test images. 

    \noindent \textbf{NuScenes Dataset.} NuScenes \cite{nuscenes} stands as a comprehensive autonomous driving dataset, featuring 3D bounding boxes for 1000 scenes. We follow the different scenes selected by S-NeRF \cite{s-nerf} and DrivingGaussian \cite{zhou2024drivinggaussian} for our experiments.


\subsection{Experimental Settings}
    \noindent \textbf{Implementation Details.} Our implementation is primarily based on Deformable 3D GS \cite{d3dgs}, maintaining consistency with its training iterations and optimizer settings. Additionally, the DEM and the OEM are optimized using Adam with a learning rate of 0.001.

    \noindent \textbf{Adaptor Model Training Settings.} Our training dataset is nuScenes poses $P$ and poses $P_{S}$ generated by SfM as data pairs. We need to train on 17 different scenes, each with approximately 230 pose pairs. Each scene requires training for 1000 epochs. We use the Adam optimizer with a learning rate of 2e-4 and a batch size of 16. It takes about 2 hours to train using AMD MI100. Adaptor model loss weights $w_{1}$, $w_{2}$ and $w_{3}$ in Equation \ref{eq:loss_all} are 50, 0.1 and 1, respectively, and the $N$ camera pose is 15.

    \noindent \textbf{2D/3D Detection Settings.} To verify the effectiveness of synthetic data, we use mainstream detection model to perform 2D and 3D detection tasks on four categories: car, bus, truck, and trailer. 2D detection tasks use Co-DETR \cite{Zong_2023_ICCV} and 3D tasks use MonoLSS \cite{li2024monolss}. For the test dataset, we select the first fifteen scene data in the nuScenes validation set.

    \begin{table}[t]
    \centering
    \adjustbox{max width=0.47\textwidth}{
    {\small
        \begin{tabular}{c|c|ccc} 
        \hline
        Methods               & Supervision                & PSNR$\uparrow$ & LPIPS $\downarrow$ & SSIM $\uparrow$           \\ 
        \hline
        NPBG                  & RGB + Points         & 19.58          & 0.248          & 0.628           \\
        ADOP                  & RGB + Points         & 20.08          & 0.183          & 0.623           \\
        Deepview              & RGB                        & 17.28          & 0.196          & 0.716           \\
        MPI                   & RGB                        & 19.54          & 0.158          & 0.733           \\
        READ                  & RGB                        & 23.48          & \textbf{0.132}          & 0.787           \\ 
        \hline
        3D GS                 & RGB                        & 20.37          & 0.357          & 0.679           \\
        Mip-splatting         & RGB                        & 20.86          & 0.341          & 0.696           \\
        Deformable 3D GS      & RGB                        & 21.33          & 0.306          & 0.713           \\ 
        \hline
        
  \textbf{EGSRAL(Ours)} & RGB                        & \textbf{23.60} & 0.192          & \textbf{0.787}           \\
        \hline
        \end{tabular}
    }
    }
    \caption{Comparisons with existing methods on KITTI City dataset. }
    \label{table:psnr_experiment}
    \end{table}

    \begin{table}[t]
    \centering
    \adjustbox{max width=0.47\textwidth}{
    {\small
        \begin{tabular}{ccccc} 
        \hline
        Methods           & Input             & PSNR$\uparrow$ & LPIPS $\downarrow$ & SSIM $\uparrow$  \\ 
        \hline
        S-NeRF            & RGB+LiDAR      & 25.43          & 0.302 & 0.730              \\
        SUDS              & RGB+LiDAR      & 21.26          & 0.466 & 0.603              \\
        EmerNeRF          & RGB+LiDAR      & 26.75          & 0.311 & 0.760              \\ 
        \hline
        3D GS             & RGB+SfM & 26.08          & 0.298 & 0.717              \\
        DrivingGaussian-L & RGB+LiDAR      & 28.74          & 0.237 & 0.865              \\ 
        \hline
         \textbf{EGSRAL(Ours)}     & RGB+LiDAR      & \textbf{29.04}          & \textbf{0.162} & \textbf{0.883}              \\
        \hline
        \end{tabular}
    }
    }
    \caption{Comparisons with existing methods on the nuScenes dataset used in DrivingGaussian.}
    \label{table:dgs}
    \end{table}

\subsection{Performance of Novel View Synthesis}
    \noindent \textbf{Evaluation on KITTI.} We compare our method with state-of-the-art (SOTA) methods, as shown in Table ~\ref{table:psnr_experiment}. Our approach demonstrates superior performance compared to other 3D GS methods, with PSNR and SSIM improvements of 2.27 and 0.074, respectively, and 0.114 reduction in LPIPS over Deformable 3D GS. Additionally, compared to other methods like READ, our method enhances the PSNR metric to 23.60 while achieving comparable SSIM results. However, the 3D GS-based method has a slight deficiency in modeling high-frequency information, such as ground texture, which leads to a higher LPIPS. For detailed visualization results, please refer to the supplementary materials.

    \begin{table}[t] \small
    \begin{center}
    \adjustbox{max width=1.0\linewidth}{
    \begin{tabular}{c|ccc}
    \hline
    Methods         & PSNR$\uparrow$    & LPIPS$\downarrow$ & SSIM$\uparrow$   \\ \hline
    Mip-NeRF        & 18.22             & 0.421             & 0.655  \\
    Mip-NeRF360     & 24.37             & 0.240             & 0.795 \\
    Urban-NeRF      & 21.49             & 0.491             & 0.661  \\
    S-NeRF          & 26.21             & 0.228             & 0.831  \\ 
    \hline
    3D GS           & 32.82             & 0.225             & 0.925  \\
    Mip-splatting   & 32.22             & 0.224             & 0.928  \\
    Deformable 3D GS & 33.43            & 0.224             & 0.932 \\ \hline
     \textbf{EGSRAL(Ours) }       & \textbf{34.43}   & \textbf{0.205}   & \textbf{0.939}     \\ \hline
    \end{tabular}
    }
    \end{center}
    \caption{Comparisons with existing methods on the nuScenes dataset used in S-NeRF.}
    \label{table:nuscence_psnr_experiment}
    \end{table}
    
    \begin{figure}[t]
        \centering 
        \includegraphics[width=0.95\linewidth]{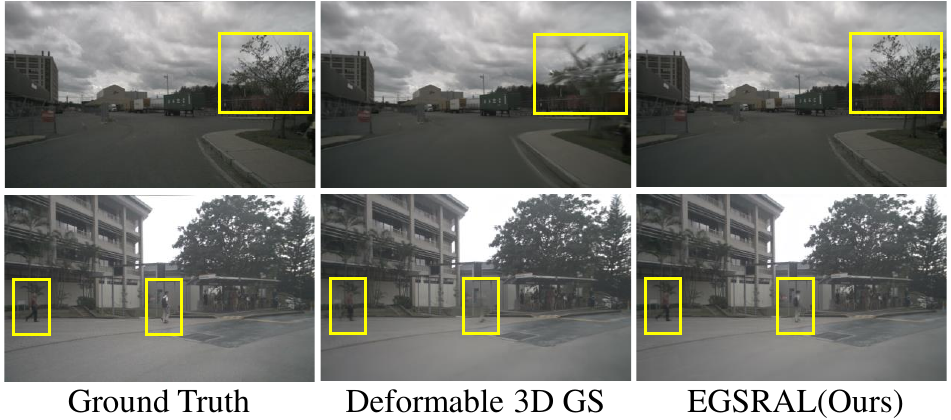}
        \caption{Qualitative comparison of novel view synthesis on the nuScenes dataset.}
        \label{fig:psnr_visual} 
    \end{figure}
    
    \noindent \textbf{Evaluation on nuScenes.} Following the scenes used by DrivingGaussian, as shown in Table \ref{table:dgs}, our method achieves the highest performance, with an increase of 0.3 and 0.018 in PSNR and SSIM, respectively, and a decrease of 0.075 in the LPIPS. It is worth noting that DrivingGaussian trains the dynamic objects and the static backgrounds separately, training an individual 3D GS model for each dynamic object then merging them. In contrast, our method requires only a single training process and does not need any annotations, resulting in better performance and reducing the dependence on data annotations.
    
    Following the scenes used by S-NeRf \cite{s-nerf}, as shown in Table \ref{table:nuscence_psnr_experiment}, we contrast our method with the SOTA method. In addition, we replicated 3D GS \cite{3dgs}, Deformable 3D GS \cite{d3dgs}, and mip-splatting \cite{Yu2024MipSplatting} on this dataset. 
    The 3D GS-based method has much higher metrics on these scenes than the NeRf-based method. Our method achieves the best performance, with PSNR improved by 1.00, SSIM improved by 0.007, and LPIPS reduced by 0.019 compared to the Deformable 3D GS.

    In Figure~\ref{fig:psnr_visual},
    we compare the view synthesis results of our method with those of other 3D GS methods. The visualization results show superior synthesis capabilities, especially achieving more realistic rendering results on static backgrounds (trees) and dynamic objects (pedestrians).

\subsection{Performance of Auto Labeling}
\begin{figure}[ht]
        \centering 
      \includegraphics[width=0.95\linewidth]{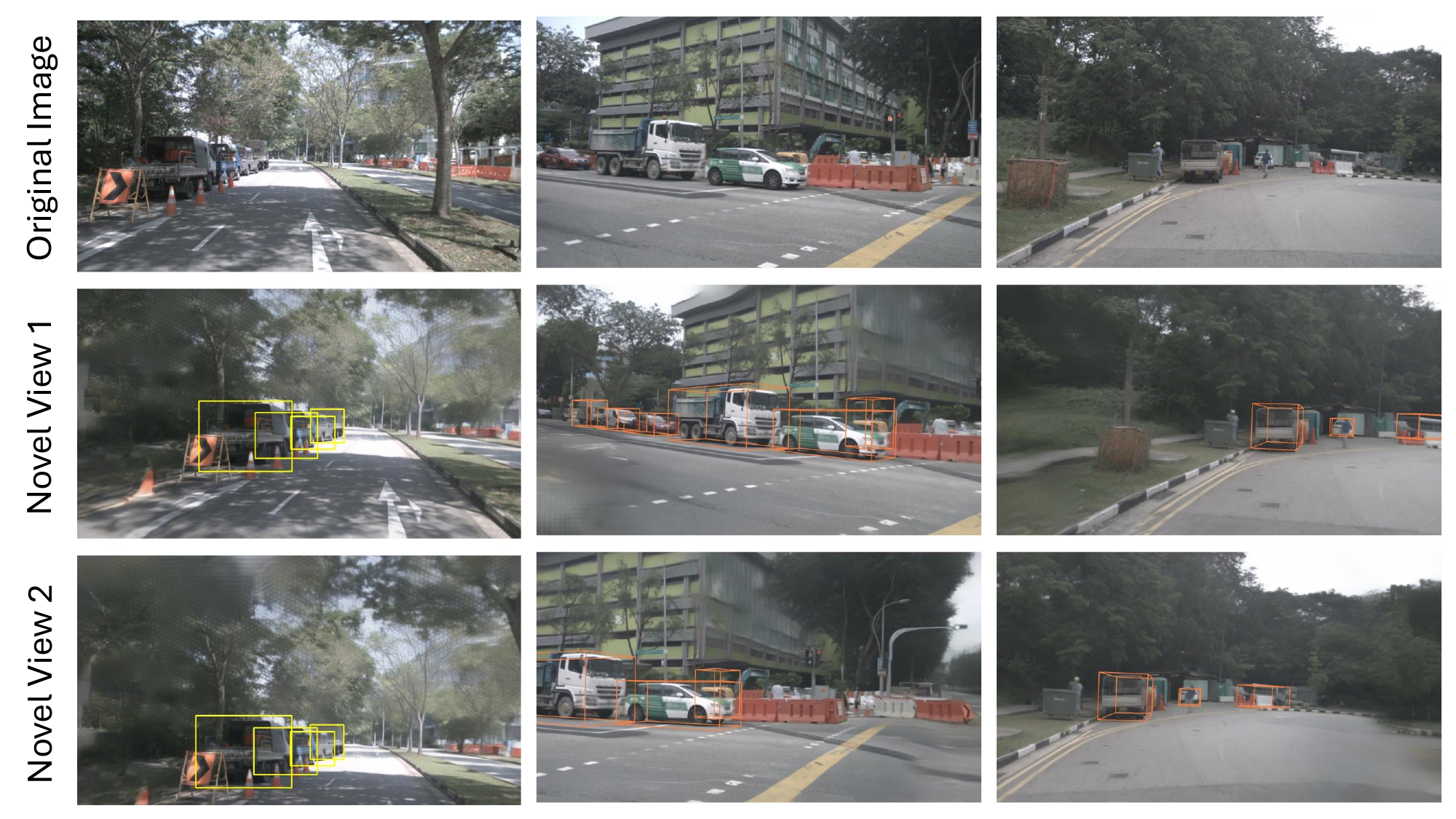}
        \caption{Visualizing 2D/3D auto labeling on nuScenes.}
        \label{fig:auto-labeling} 
    \end{figure}

    We select 17 different scenes from the nuScenes dataset for the 2D/3D auto-labeling task. Using the adaptor, we generate novel view images and automatically label them simultaneously. The results are presented in Figure~\ref{fig:auto-labeling}. The three columns represent three different scenes: the first row displays the original images, while the last two rows show the novel view images with auto labeling. In these images, the yellow boxes denote 2D annotations, and the orange boxes denote 3D annotations. The results demonstrate that our adaptor performs effectively across different scenes.
    
\noindent \textbf{Performance of 2D/3D Object Detection.}
    We select 17 different scenes containing 3,898 images (674 of which are sample images with annotations) from the nuScenes dataset as the baseline dataset for the detection task. We produce the annotations for the non-sample images based on the annotations of sample images by camera coordinate system transformation. 
    To augment the dataset, we synthesize additional images using an adaptor, increasing the total number of images to two and three times that of the baseline dataset. The 2D detection results are presented in Table \ref{table:2ddetection_experiment}, and the 3D detection results are presented in Table \ref{table:3ddetection_experiment}. In these tables, $sample$ refers to the 674 sample images, while $all$ refers to the 3,898 images, including both sample and non-sample images. The results demonstrate that, without any additional input, our data augmentation method improves the accuracy of the detection model in both the $sample$ and the $all$ dataset.

    \begin{table}[t] \small
    \begin{center}
    \adjustbox{max width=1.0\linewidth}{
    {\small
        \begin{tabular}{c|c|c|ccc} 
        \hline
        Model  & Dataset       & Total\_amount & mAP & AP50 & AP75 \\ \hline
        \multirow{6}{*}{Co-DETR}  & \multirow{3}{*}{sample}         & 1$\times$            & 23.7   &  44.6   & 22.9    \\ 
                                &                                   & 2$\times$            & 25.9   &  45.9    & 23.1     \\
                                &                                    &  3$\times$            & \textbf{26.8}    &  \textbf{46.1}    & \textbf{23.2}     \\ \cline{2-6}
                               & \multirow{3}{*}{all}    & 1$\times$           & 28.1   &  46.5   & 24.2   \\ 
                                &      & 2$\times$            & 29.9   &  47.7    & 24.8  \\
                                &      &  3$\times$            &  \textbf{31.3}    &   \textbf{48.2}    &  \textbf{25.3}     \\  \hline                                                  
        \end{tabular}
        }
        }
        \caption{Augmentation results of 2D detection on nuScenes.}
        \label{table:2ddetection_experiment}
    \end{center}
    \end{table}

    \begin{table}[t] \small
    \begin{center}
    \adjustbox{max width=1.0\linewidth}{
    {\small
        \begin{tabular}{c|c|c|ccc} 
        \hline
        Model  & Dataset       & Total\_amount & AP3d & Mod. & Hard \\ \hline
        \multirow{6}{*}{MonoLSS}  & \multirow{3}{*}{sample}         & 1$\times$            & 17.21    &  12.55   & 11.30    \\ 
                                &                                   & 2$\times$            & 19.78    & 13.32    & 12.81     \\
                                &                                    &  3$\times$            &  \textbf{20.31}    &   \textbf{13.85}  &  \textbf{12.97}     \\ \cline{2-6}
                               & \multirow{3}{*}{all}    & 1$\times$            & 21.69    &  14.27   & 13.95  \\ 
                                &      & 2$\times$            & 22.15    &  14.93    & 14.22   \\
                                &      &  3$\times$            &  \textbf{22.87}   &   \textbf{15.25}    &  \textbf{14.40} \\  \hline                                                
        \end{tabular}
        }
        }
        \caption{Augmentation results of 3D detection on nuScenes.}
        \label{table:3ddetection_experiment}
    \end{center}
    \end{table}

\subsection{Ablation Study and Analysis}
    In this section, we provide ablation experiments for different modules of our method on the KITTI City dataset, as shown in Table ~\ref{table:ablation_experiment}. For more experiments and analyses, please see Appendix Section 3. Comparing rows 2 and 3, we observe that using the DEM alone to refine the deformation field improves PSNR by 1.04 and SSIM by 0.03. In contrast, using the OEM alone results in improvements of 0.99 in PSNR and 0.025 in SSIM, with a reduction of 0.029 in LPIPS. When both DEM and OEM are combined, the evaluation metrics show significant improvements over Deformable 3D GS, demonstrating the enhanced synthesis capabilities of our method for large-scale scene modeling. Additionally, we validate our proposed grouping strategy. As shown in the last row, our approach further enhances PSNR by 0.76, improves SSIM by 0.027, and reduces LPIPS by 0.061 by incorporating the grouping strategy. In conclusion, the above ablation studies fully validate the effectiveness of all proposed modules in our method.



    \begin{table}[!t] \small
    \begin{center}
    \adjustbox{max width=1.0\linewidth}{
    \begin{tabular}{ccc|ccc} 
    \hline
    DEM & OEM & GPS & PSNR$\uparrow$ & LPIPS$\downarrow$ & SSIM$\uparrow$  \\ 
    \hline
    -       & -       & -        & 21.33  & 0.306   & 0.713   \\
    $\checkmark$       & -       & -        & 22.37  & 0.279   & 0.743   \\
    -       & $\checkmark$       & -        & 22.32  & 0.277   & 0.738   \\
    $\checkmark$       & $\checkmark$       & -        & 22.84  & 0.253   & 0.760   \\ 
    $\checkmark$       & $\checkmark$       & $\checkmark$        & \textbf{23.60}  & \textbf{0.192}   & \textbf{0.787}   \\
    \hline
    \end{tabular}
    }
    \end{center}
    \caption{Effect of each module in our method.}
    \label{table:ablation_experiment}
    \end{table}
    

\section{Conclusion}
    In this paper, we present EGSRAL, a novel 3D GS-based renderer with an automated labeling framework capable of synthesizing novel view images with corresponding annotations.
    For novel view rendering, we introduce two effective modules to improve the ability of 3D GS to model complex scenes and propose a grouping strategy to solve the problem of unreasonable view of large-scale scenes. For novel view auto labeling, we propose an adaptor to generate new annotations for novel views. Experimental results show that EGSRAL significantly outperforms existing methods in novel view synthesis and achieves superior object detection performance on annotated images.

\bibliography{aaai25}

\section{Appendix}
In this supplementary material, we present more implementation details and additional experimental results.
\begin{itemize}
    \item In Section \ref{sec: implementation details}, we present the training details of our method, the specific implementation of the grouping strategy algorithm, and the performance comparison of the latest 2D and 3D detectors.
    \item In Section \ref{sec: additional results}, we present additional experimental results.
    \item In Section \ref{sec: visualization}, we present the KITTI dataset visualization results and more adaptor visualization results.
\end{itemize}

\section{Implementation Details}
\label{sec: implementation details}
\subsection{Experiment Setting}
\label{sec: setting}
\noindent\textbf{KITTI Selected by READ.}
    For the KITTI dataset, we follow the setup of READ \cite{read} and use the single camera as input, with a resolution of $1232 \times 368$. We select every 10th image of cameras in the sequences as the test set. See Table \ref{table: hyperparameters} for the detailed hyperparameter settings.
    
\noindent\textbf{NuScenes-S Selected by S-NeRF.}
    For the NuScenes-S dataset, we follow the setup of S-NeRF \cite{s-nerf} and only use the front camera as input, with a resolution of $1600 \times 900$. We select every 4th image of cameras in the sequences as the test set. The specific scene tokens are 164, 209, 359, and 916. We report the average results of all camera frames on the selected scenes and assess our models using the average score of PSNR, SSIM, and LPIPS. See Table \ref{table: hyperparameters} for the detailed hyperparameter settings.
    
\noindent\textbf{NuScenes-D Selected by DrivingGaussian.}

\begin{table*}[ht] \small
    \begin{center}
    \adjustbox{max width=1.0\textwidth}{
    {\small
        \begin{tabular}{c|cc|cc|cc|ccc} 
        \hline
        \multirow{2}{*}{Hyperparameter} & \multicolumn{2}{c|}{3DGS} & \multicolumn{2}{c|}{Mip-Splatting} & \multicolumn{2}{c|}{D3DGS} & \multicolumn{3}{c}{EGSRAL (Ours)}  \\ 
        \cline{2-10}
                                        & KITTI     & nuScenes-S    & KITTI     & nuScenes-S             & KITTI     & nuScenes-S     & KITTI  & nuScenes-S & nuScenes-D   \\ 
        \hline
        Training iterations             & 2,000,000 & 40,000        & 2,000,000 & 40,000                 & 2,000,000 & 40,000         & 40,000 & 40,000     & 200,000      \\
        Warm up iteration                        & -         & -             & -         & -                      & 90,000    & 3,000          & 3,000  & 3,000      & 30,000       \\
        Densification interval          & 3,000     & 100           & 3,000     & 100                    & 3,000     & 100            & 100    & 100        & 300          \\
        Opacity reset interval          & 90,000    & 3,000         & 90,000    & 3,000                  & 90,000    & 3,000          & 3,000  & 3,000      & 6,000        \\
        Densify from iteration          & 15,000    & 500           & 15,000    & 500                    & 15,000    & 500            & 500    & 500        & 5,000        \\
        Densify until iteration         & 500,000 & 15,000        & 500,000 & 15,000                 & 500,000 & 15,000         & 15,000 & 15,000     & 100,000      \\
        Densify grad threshold          & 0.0002    & 0.0002        & 0.0002    & 0.0002                 & 0.0002    & 0.0002         & 0.0002 & 0.0002     & 0.0001       \\
        \hline
        \end{tabular}
        }
        }
        \caption{Hyperparameter configuration for different datasets. The nuScenes-S denotes the dataset selected by the S-NeRF, and the nuScenes-D denotes the dataset selected by the DrivingGaussian.}
        \label{table: hyperparameters}
    \end{center}
    \end{table*}
    For the NuScenes-D dataset, we follow the setup of DrivingGaussian \cite{zhou2024drivinggaussian} and use synchronized images from 6 cameras in surrounding views as inputs, with a resolution of $1600 \times 900$. We select every 5th image of cameras in the sequences as the test set. The specific scene tokens are 103, 168, 212, 220, 228, and 687. We report the average results of all camera frames on the selected scenes and assess our models using the average score of PSNR, SSIM, and LPIPS. See Table \ref{table: hyperparameters} for the detailed hyperparameter settings.

    To compare with DrivingGaussian, we utilize LiDAR point clouds to initialize the 3D Gaussians. Specifically, we stack the LiDAR point clouds of the entire scene and retrieve the corresponding image color of each point through the projection relationship. Finally, the voxelized downsampled point cloud is used to initialize the 3D Gaussians.
    
\subsection{The Detailed Grouping Strategy}
\label{sec: gps}
    \begin{algorithm}[t]
    \small
    \SetAlgoLined
    \SetKwInOut{Input}{Input}
    \SetKwInOut{Output}{Output}
        \Input{Image sequences $\boldsymbol{I} \in R^{N_I \times H \times W}$. \\
            Corresponding 3D Gaussians $\boldsymbol{G}$. \\
            Number of image per group $N_{per}$. \\
            Number of group number $N_{g} = \lceil N_{I} / N_{per} \rceil$. \\
            Valid distance of the image $d$. \\
            Number of training iterations $T$. \\
            Number of overlap training $N_o$. \\
            }

        {\textbf{1. Grouping:}} \\
        \tcp*[f]{Group images and 3D Gaussians}. \\
        {$G_{img} = \text{List} \ [\ ]$} \\
        \For{$i=1:N_{g}$}
            {
                {{$G_{img}[i]=\boldsymbol{I}[i \times N_{per} : (i+1) \times N_{per}]$};\\
                {$mask=\left\{\text{False}\right\}^{|\boldsymbol{G}|\times 1}$}}
                
                \For{$j=1:N_{per}$}
                    {
                        {$mask = mask \ | \  (dist(\boldsymbol{G}.xyz, G_{img}[i][j].xyz) < d)$}
                    }
                {\tcp{Assign group id to 3D Gaussians.}}
                {$\boldsymbol{G}[mask].id = i$}
            }
            
        {\textbf{2. Grouping for Training:}} \\
        \For{$i=1:T$}
            {
                \For{$j=1:N_g$}
                {
                    \If{$j \ne 1$}
                    {
                        \tcp{Overlap Training.}
                        $img_{gt} = \text{random}(G_{img}[j] + G_{img}[j-1][-N_{o}:])$
                    }
                    \Else
                    {
                        $img_{gt} = \text{random}(G_{img}[j])$
                    }
                    {$img_{out} = EGSRAL(\boldsymbol{G}[\boldsymbol{G}.id==j], img_{gt}.view)$ \\
                    $loss = Loss(img_{out}, img_{gt})$ \\
                    $loss.backward()$}
                }
                {$EGSRAL.Optimizer.step()$}
            }
        {$EGSRAL.save()$}
    \caption{Grouping Strategy}
    \label{alg: GPS}
    \end{algorithm}

    We propose the grouping strategy for the large-scale driving scene reconstruction. While DrivingGaussian \cite{zhou2024drivinggaussian} also mentions incremental training, this is for the small nuScenes scene and does not carefully consider how to apply it to large-scale driving scenes like KITTI.
    Table \ref{table: drivinggaussian vs ours} shows the difference between our method and DrivingGaussian, although both of them group the scene point cloud, DrivingGaussian is static incremental training, which needs to be trained in the order of the groups. On the contrary, our method only needs to assign a group $id$ to each Gaussian primitive, and the training process is shuffled. In addition, the DrivingGaussian method needs to train a model for each dynamic object and grouping scene during training. This approach not only depends on additional annotations but also necessitates the training of multiple models. In contrast, our method eliminates the need for extra annotations and requires training only a single model.
    
\begin{figure}[t]
    \centering 
    \includegraphics[width=1.0\linewidth]{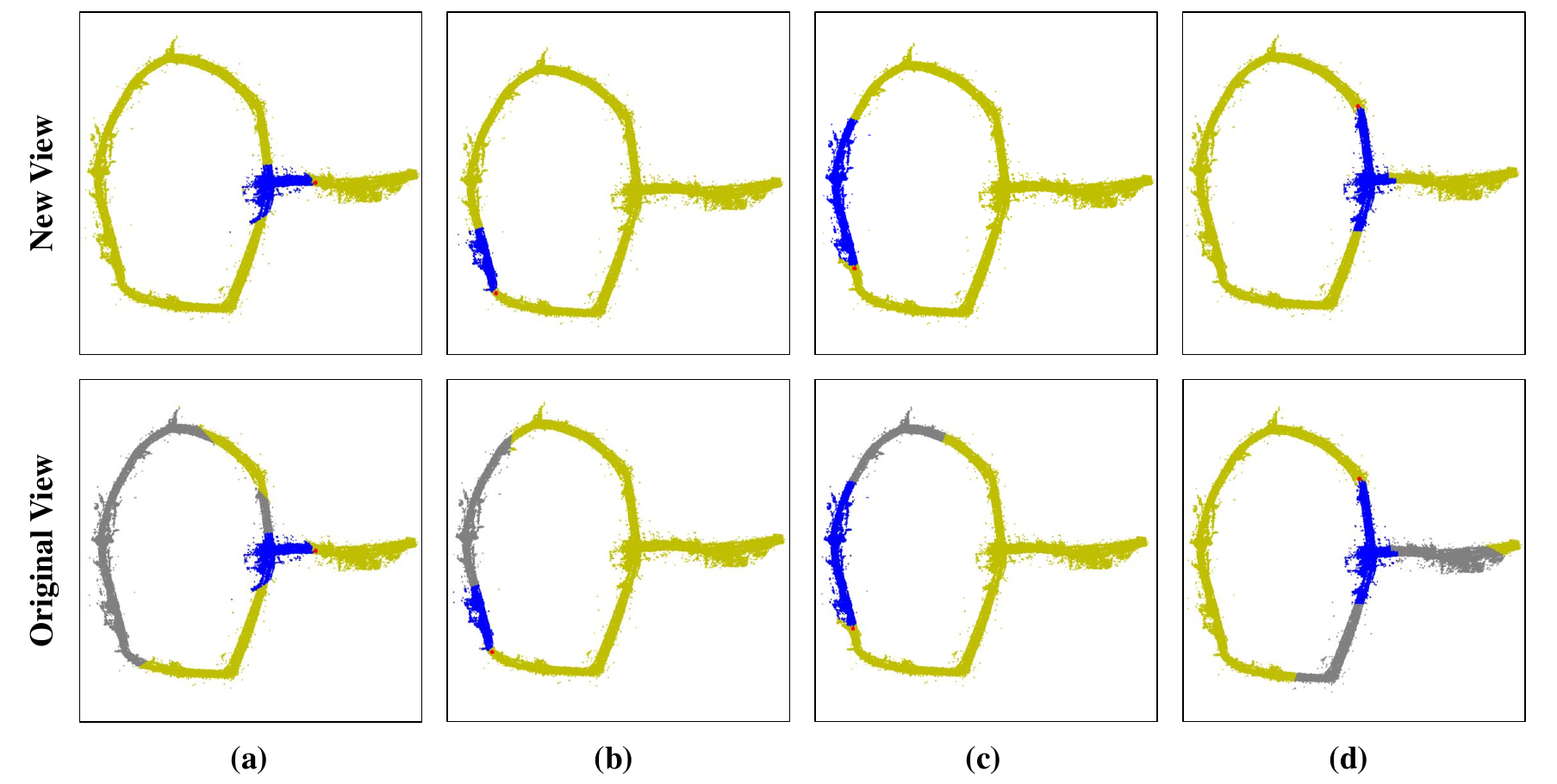}
    \caption{Field of view comparison of large-scale scenes. The new view is the group results by the grouping strategy, and the original view is the original range of the view, where the blue points indicate points that participate in rendering, yellow points that do not participate in rendering, and gray points are not included in the new field of view.}
    \label{fig: gps_2} 
    \end{figure}

    Algorithm \ref{alg: GPS} presents the detailed pseudocode implementation, while Figure \ref{fig: gps_2} provides additional visualizations of the grouping strategy. By utilizing this strategy, we not only achieve more efficient reconstruction of large-scale driving scenes but also address issues related to unreasonable fields of view.
    
    \begin{table}[t] \small
    \begin{center}
    \adjustbox{max width=0.5\textwidth}{
    {\small
        \begin{tabular}{c|c|c} 
        \hline
                 & DrivingGaussian & EGSRAL(Ours)  \\ 
        \hline
        Group         & $\checkmark$    & $\checkmark$             \\
        Training mode & Order           & Shuffle       \\
        Network Number  & $N_g + N_{obj} + 1$ & 1             \\
        \hline
        \end{tabular}
        }
        }
        \caption{Comparison between our grouping strategy and DrivingGaussian's incremental training. The $N_g$ is the number of groups, and the $N_{obj}$ denotes the number of dynamic objects.}
        \label{table: drivinggaussian vs ours}
    \end{center}
    \end{table}
    
    \begin{table}[t]
        \centering
        \adjustbox{max width=1.0\linewidth}{
        {\small
        \begin{tabular}{cccc} 
        \hline
        Detector     & Year      & Backbone &  AP   \\ 
        \hline
        MS-DETR    & CVPR 2024 & R50      & 51.7  \\
        RT-DETR    & CVPR 2024 & R50      & 53.1  \\
        Co-DETR    & ICCV 2023 & R50      & 54.8  \\
        DINO       & ICLR 2023 & R50      & 49.4  \\
        Group DETR & ICCV 2023 & R50      & 50.1  \\
        \hline
        \end{tabular}}
        }
        \caption{2D detector performance comparison.}
        \label{table:2d dector compare}
    \end{table}

\subsection{SOTA 2D/3D Detector}
    \textbf{2D Detector.} We survey recent works on 2D detection, as shown in Table \ref{table:2d dector compare}. Co-DETR \cite{zong2023codetr} achieves high AP on the COCO validation set, demonstrating strong competitiveness compared to several current mainstream methods. This performance positions it as a SOTA model in object detection.

    \noindent\textbf{3D Detector.} MonoLss \cite{li2024monolss} achieves state-of-the-art performance in the primary categories of car, cyclist, and pedestrian on the KITTI monocular 3D object detection benchmark, as shown in Table \ref{table:3d dector compare}. It also demonstrates strong competitiveness in cross-dataset evaluations on the Waymo, KITTI, and nuScenes datasets. At the time of our experiments, MonoLss is the leading open-source 3D detection model that relies solely on monocular image input.

    \begin{table}
    \centering
    \adjustbox{max width=1.0\linewidth}{
    {\small
    \begin{tabular}{ccccc} 
    \hline
    \multirow{2}{*}{Detector} & \multirow{2}{*}{Year} & \multicolumn{3}{c}{AP\_BEV (IoU=0.7\textbar{}R40)}  \\ 
    \cline{3-5}
                              &                       & Easy  & Mod.  & Hard                                \\ 
    \hline
    MonoLss                   & 3DV 2024              & 34.89 & 25.95 & 22.59                               \\
    DEVIANT                   & ECCV 2022             & 29.65 & 20.44 & 17.43                               \\
    MonoCon                   & AAAI 2022             & 31.12 & 22.10 & 19.00                               \\
    MonoDDE                   & CVPR 2022             & 33.58 & 23.46 & 20.37                               \\
    \hline
    \end{tabular}}
    }
    \caption{3D detector performance comparison.}
    \label{table:3d dector compare}
    \end{table}

    \begin{table*}[ht] \small
    \begin{center}
    \adjustbox{max width=1.0\textwidth}{
    {\small
        \begin{tabular}{c|ccc|ccc|ccc|ccc|ccc} 
        \hline
        \multirow{2}{*}{Method} & \multicolumn{3}{c|}{164}                            & \multicolumn{3}{c|}{209}                            & \multicolumn{3}{c|}{359}                            & \multicolumn{3}{c|}{916}                            & \multicolumn{3}{c}{Average}                          \\ 
        \cline{2-16}
                                & PSNR$\uparrow$ & SSIM$\uparrow$ & LPIPS$\downarrow$ & PSNR$\uparrow$ & SSIM$\uparrow$ & LPIPS$\downarrow$ & PSNR$\uparrow$ & SSIM$\uparrow$ & LPIPS$\downarrow$ & PSNR$\uparrow$ & SSIM$\uparrow$ & LPIPS$\downarrow$ & PSNR$\uparrow$ & SSIM$\uparrow$ & LPIPS$\downarrow$  \\ 
        \hline
        3DGS                    & 35.45          & 0.942          & 0.209             & 33.71          & 0.950          & 0.248             & 30.78          & 0.903          & 0.240             & 31.32          & 0.926          & 0.201             & 32.82          & 0.925          & 0.225              \\
        Mip-Splatting           & 34.28          & 0.935          & 0.213             & 32.84          & 0.944          & 0.252             & 30.41          & 0.901          & 0.238             & 31.34          & 0.931          & 0.191             & 32.22          & 0.928          & 0.224              \\
        Deformable 3D GS        & 35.87          & 0.943          & 0.208             & 35.37          & 0.956          & 0.241             & 30.48          & 0.898          & 0.246             & 32.00          & 0.930          & 0.199             & 33.43          & 0.932          & 0.224              \\
        \hline
        \textbf{EGSRAL(Ours)}   & \textbf{36.32} & \textbf{0.946} & \textbf{0.193}    & \textbf{36.48} & \textbf{0.963} & \textbf{0.216}    & \textbf{31.94} & \textbf{0.911} & \textbf{0.218}    & \textbf{32.99} & \textbf{0.936} & \textbf{0.191}    & \textbf{34.43} & \textbf{0.939} & \textbf{0.205}     \\
        \hline
        \end{tabular}
        }
        }
        \caption{Detailed experimental results for different scenes on the nuScenes-S dataset.}
        \label{table: nus-s}
    \end{center}
    \end{table*}

    \begin{table*}[t] \small
    \begin{center}
    \adjustbox{max width=1.0\textwidth}{
    {\small
    \begin{tabular}{c|c|cccccc|c} 
    \hline
    Method                           & Metric & 103   & 168   & 212   & 220   & 228   & 687   & Avg.            \\ 
    \hline
    \multirow{3}{*}{\textbf{EGSRAL(Ours)}}    & PSNR$\uparrow$   & 28.87 & 28.52 & 30.10 & 28.87 & 29.36 & 28.48 & \textbf{29.04}  \\
                                     & SSIM$\uparrow$   & 0.874 & 0.884 & 0.905 & 0.863 & 0.889 & 0.885 & \textbf{0.883}  \\
                                     & LPIPS$\downarrow$  & 0.170 & 0.168 & 0.141 & 0.178 & 0.159 & 0.153 & \textbf{0.162}  \\ 
    \hline
    \multirow{3}{*}{DrivingGaussian} & PSNR$\uparrow$   & -     & -     & -     & -     & -     & -     & 28.74           \\
                                     & SSIM$\uparrow$   & -     & -     & -     & -     & -     & -     & 0.865           \\
                                     & LPIPS$\downarrow$  & -     & -     & -     & -     & -     & -     & 0.237           \\
    \hline
    \end{tabular}
        }
        }
        \caption{Detailed experimental results for different scenes on the nuScenes-D dataset.}
        \label{table: nus-d}
        \vspace{-4mm}
    \end{center}
    \end{table*}

\section{Additional Experiments}
\label{sec: additional results}
    \subsection{Detailed Results on nuScenes} 
    In Tables \ref{table: nus-s} and \ref{table: nus-d} we provide specific metrics for each scene that we detail on nuScenes-S and nuScenes-D dataset, respectively, where we achieve the best performance compared to other SOTA methods.

    \subsection{Ablation Study}
    \noindent \textbf{Effect of the Group Number in GPS.} In Table \ref{table: gps},  we present the results of our grouping strategy under different numbers of groups. With a small number of groups, the large scene leads to an unreasonable field of view after grouping, preventing effective training for each group. Conversely, with too many groups, the correct field of view is restricted, increasing the optimization burden of 3D GS. Ultimately, we achieve the best performance with 8 groups.

    \begin{table}[t] \small
    \begin{center}
    \adjustbox{max width=1.0\linewidth}{
    \begin{tabular}{c|cccc} 
    \hline
    Group & 4     & 6     & 8     & 10     \\ 
    \hline
    PSNR$\uparrow$  & 22.64 & 23.10 & \textbf{23.60} & 23.38  \\
    \hline
    \end{tabular}
    }
    \end{center}
    \caption{Effect of the group number in GPS.}
    \label{table: gps}
    \end{table}

    \noindent \textbf{Effect of the Number of Camera Poses.}    
    For the adaptor, we conduct an ablation study on the number of camera poses, as shown in Table~\ref{table:3ddetection_ablation}. During the adaptor's training process, the number of camera poses $N$  is set to 15. By comparing the 2nd, 3rd, and 4th rows, it is evident that, compared to using 10 or 20 camera poses, selecting 15 has improved the accuracy of 3D frame alignment in novel-view images, enhancing the performance of the 3D detection task.

    \begin{table}[th] \small
    \begin{center}
    \adjustbox{max width=1.0\linewidth}{
    {\small
        \begin{tabular}{c|c|ccc} 
        \hline
        Model         & $N$  & AP3D & Mod. & Hard \\ 
        \hline
        \multirow{4}{*}{MonoLSS} &      
        -             & 17.21     & 12.55     & 11.30    \\
        &            10            & 18.55     & 12.87     & 12.11     \\
         &                         
        \textbf{15}      &  \textbf{19.78}     &  \textbf{13.32}     & \textbf{12.81}     \\
        & 20            & 19.67     & 13.26     & 12.73     \\                  
        \hline                                                  
        \end{tabular}
        }
        }
        \caption{Effect of the number of camera poses in adaptor.}
        \label{table:3ddetection_ablation}
        \vspace{-5mm}
    \end{center}
    \end{table}
    \subsection{Necessity Analysis of Adaptor}
    To address the complex alignment errors among camera poses in OWCS and EWCS, we propose a trainable adaptor with sampling pose augmentation, replacing the traditional transformation matrix estimation using the Umeyama algorithm \cite{umeyama1991least}. The SfM method used in 3D GS is monocular, which limits its ability to produce high-quality reconstructions across diverse scenes. This limitation often results in the introduction of outliers, such as matching errors within the SfM process, leading to inconsistent alignment. These challenges make it difficult for traditional methods to achieve reliable results in such scenarios.

    \begin{table}[th] \small
    \begin{center}
    \adjustbox{max width=1.0\linewidth}{
    {\small
        \begin{tabular}{c|cc}
        \hline
        Metric & Matrix & Adaptor \\ \hline
        AP (\%)     & 21.71     & \textbf{72.53}      \\
        AD ($m$)    & 1.867     & \textbf{0.605}      \\ \hline
        \end{tabular}
        }
        }
        \caption{Comparison of AP and average 3D center distance between GT and generated 3D boxes in OWCS.}
        \label{table:adaptor_com}
    \vspace{-4mm}
    \end{center}
    \end{table}
    \begin{figure}[b]
    \centering 
    \includegraphics[width=1.0\linewidth]{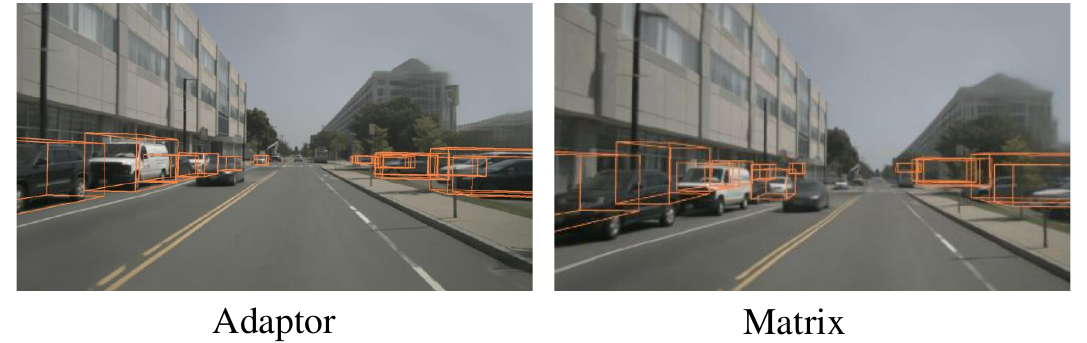}
    \caption{3D boxes projection quality comparison.}
    \label{fig: adaptor_compare} 
    \end{figure}
    To evaluate their accuracy, we conduct experiments on the nuScenes-655 scene, reserving one-fourth of the data as a test set and using the remaining data to optimize both methods. During testing, the camera poses in EWCS are first transformed into corresponding poses in OWCS via both methods. Then the fixed 3D positions of objects in the vehicle coordinate system are transformed into global 3D positions in OWCS based on the generated camera poses of both methods, and the new 3D annotation positions in OWCS of each transformation method can be obtained. We calculate the AP and average 3D center distance between the GT 3D boxes and the generated 3D boxes in OWCS, as shown in Table \ref{table:adaptor_com}. Our method significantly outperforms the standard matrix transformation-based method in both AP and average distance (AD) metrics, highlighting the advantages of our adaptor. Additionally, the visualization results in Figure \ref{fig: adaptor_compare} further demonstrate the superiority of our method.

    To verify the robustness of the Adaptor's data augmentation method, we expanded the original 2D/3D object detection dataset by doubling its size—from 17 scenes containing 3,898 images (674 of which are annotated sample images) to 34 scenes containing 8,497 images (1,455 of which are annotated sample images). As shown in Tables \ref{table:2ddetection_experiment_2} and \ref{table:3ddetection_experiment_2}, the Adaptor's data augmentation method consistently demonstrates significant improvements in the accuracy of 2D/3D object detection.

    \begin{table}[t] \small
    \begin{center}
    \adjustbox{max width=1.0\linewidth}{
    {\small
        \begin{tabular}{c|c|c|ccc} 
        \hline
        Model  & Dataset       & Total\_amount & mAP & AP50 & AP75 \\ \hline
        \multirow{6}{*}{Co-DETR}  & \multirow{3}{*}{sample}         & 1$\times$            & 28.6   &  47.1   & 24.5    \\ 
                                &                                   & 2$\times$            & 30.2   &  47.8    & 25.1     \\
                                &                                    &  3$\times$            & \textbf{31.5}    &  \textbf{48.4}    & \textbf{25.4}     \\ \cline{2-6}
                               & \multirow{3}{*}{all}    & 1$\times$           & 32.4   &  49.0   & 25.8   \\ 
                                &      & 2$\times$            & 33.2   &  49.5    & 25.9  \\
                                &      &  3$\times$            &  \textbf{33.7}    &   \textbf{49.7}    &  \textbf{26.2}     \\  \hline                                                  
        \end{tabular}
        }
        }
        \caption{Augmentation results of 2D detection on nuScenes.}
        \label{table:2ddetection_experiment_2}
        \vspace{-6mm}
    \end{center}
    \end{table}

    
    \begin{table}[b] \small
    \begin{center}
    \vspace{-4mm}
    \adjustbox{max width=1.0\linewidth}{
    {\small
        \begin{tabular}{c|c|c|ccc} 
        \hline
        Model  & Dataset       & Total\_amount & AP3d & Mod. & Hard \\ \hline
        \multirow{6}{*}{MonoLSS}  & \multirow{3}{*}{sample}         & 1$\times$            & 22.16    &  14.89   & 14.10    \\ 
                                &                                   & 2$\times$            & 23.36    & 15.36    & 14.55     \\
                                &                                    &  3$\times$            &  \textbf{23.89}    &   \textbf{15.61}  &  \textbf{14.86}     \\ \cline{2-6}
                               & \multirow{3}{*}{all}    & 1$\times$            & 24.26    &  15.83   & 15.02  \\ 
                                &      & 2$\times$            & 24.73    &  16.05    & 15.19   \\
                                &      &  3$\times$            &  \textbf{24.95}   &   \textbf{16.19}    &  \textbf{15.25} \\  \hline                                                
        \end{tabular}
        }
        }
        \caption{Augmentation results of 3D detection on nuScenes.}
        \label{table:3ddetection_experiment_2}
    \end{center}
    \end{table}
    
\subsection{Complexity and Limitations}
    \begin{figure*}[ht]
    \centering 
    \includegraphics[width=1.0\linewidth]{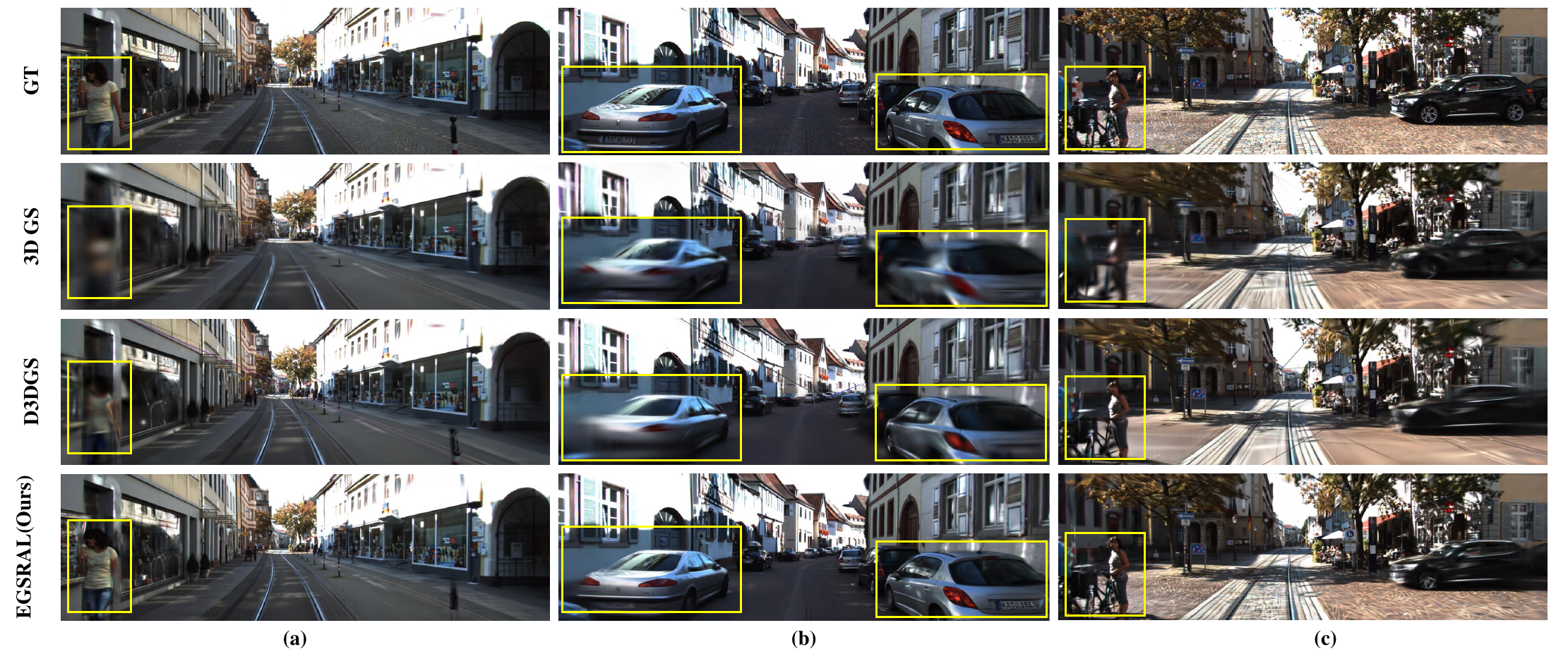}
    \vspace{-5mm}
    \caption{Qualitative comparison of novel view synthesis on the KITTI City dataset.}
    \label{fig: kitti_vis}
    \vspace{-3mm}
    \end{figure*}
    \noindent\textbf{Complexity.} Our rendering model builds upon and improves Deformable 3D GS. To demonstrate the effectiveness of our approach, we compare the average rendering speed between Deformable 3D GS and our method on a reconstructed KITTI scene. Specifically, on an NVIDIA V100 with approximately 4 million Gaussian primitives, Deformable 3D GS requires 0.768s to render a single frame, whereas our method (with 8 groups) only takes 0.254s. By applying our grouping strategy to Deformable 3D GS, the rendering speed is further improved to 0.199s. These results highlight that, despite the additional computational complexity introduced by our two enhancement modules, our method still achieves superior rendering speed compared to Deformable 3D GS. This improvement is primarily attributed to our novel grouping strategy, which efficiently groups Gaussian primitives and mitigates issues related to an excessive field of view. As a result, the number of Gaussian primitives required for rendering is reduced, significantly accelerating the overall rendering process.

    \noindent\textbf{Limitations.} Since our rendering method is based on the 3D GS framework, it inherits the limitations of this approach. The 3D GS process begins by initializing the attributes of the 3D GS primitives—such as center position, opacity, and 3D covariance matrix—using point clouds generated by the SfM method. This is followed by an optimization and training phase. Consequently, the quality of the reconstruction depends on the accuracy of the point clouds produced by the SfM method.

\section{Visualization}
\label{sec: visualization}

\subsection{Visualization on KITTI.}
    Figure \ref{fig: kitti_vis} shows the rendering results of the 3D GS-based method. It can be easily seen that our method obtains the best rendering quality among the 3D GS-based methods. However, a closer inspection shows that the 3D GS-based method cannot render the high-frequency information of the ground texture normally, which is also an important reason for its low LPIPS metric.

\subsection{More Adaptor Auto Labeling Result.}
\begin{figure}[t]
    \centering 
    \includegraphics[width=1.0\linewidth]{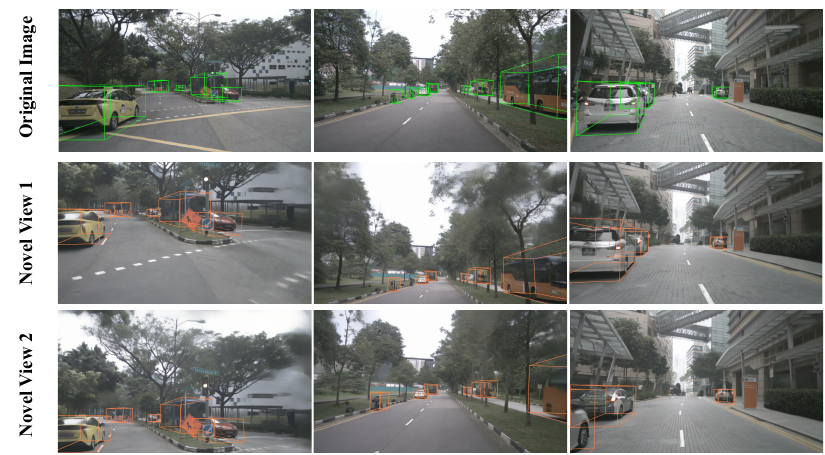}
    \caption{3D auto labeling result.}
    \label{fig: autolabelling_3d} 
    \vspace{-4mm}
    \end{figure}

\begin{figure}[bh]
    \centering 
    \includegraphics[width=1.0\linewidth]{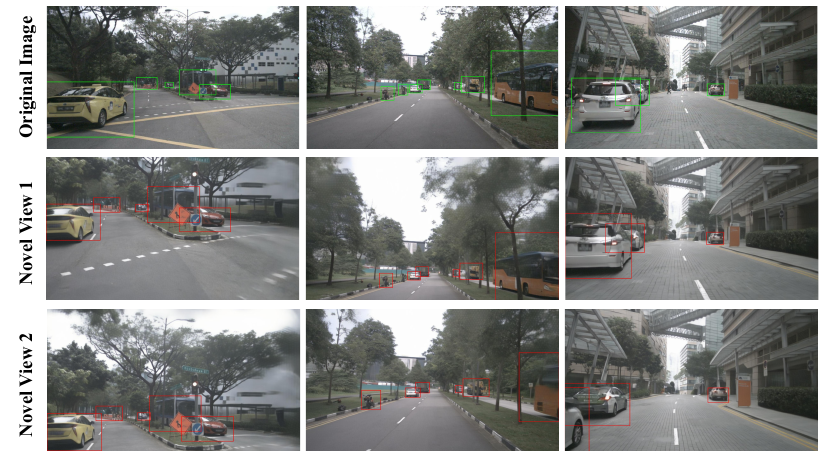}
    \vspace{-2mm}
    \caption{2D auto labeling result.}
    \label{fig: autolabelling_2d} 
    \vspace{-3mm}
    \end{figure}

    We provide more qualitative results of auto labeling via our adaptor.
    The 3D results are presented in Figure \ref{fig: autolabelling_3d}, and the 2D results are shown in Figure \ref{fig: autolabelling_2d}. The three columns represent three different scenes. The first row displays the original images and their annotations, indicated by a green frame, while the last two rows show the novel view images with auto-labeling generated by the adaptor. The results demonstrate that our adaptor performs effectively across different scenes in both 2D and 3D annotations.

\end{document}